%% file: main.tex
\documentclass[a4paper,margin=1in]{article}
\usepackage{geometry}
\usepackage[toc,page]{appendix}
\usepackage{tikz}
\usepackage{hyperref}
\usepackage{amsmath}
\usepackage{amsfonts}
\usetikzlibrary{shapes}
\usepackage{xspace}

\begin{document}



\title{RecSys Challenge 2016: job recommendations based on preselection of offers and gradient boosting}
\newcommand{\footremember}[2]{%
    \footnote{#2}
    \newcounter{#1}
    \setcounter{#1}{\value{footnote}}%
}
\newcommand{\footrecall}[1]{%
    \footnotemark[\value{#1}]%
} 
\author{Andrzej Pacuk\footremember{MIMUW}{Institute of Informatics, University
of Warsaw, Poland} \and Piotr Sankowski\footrecall{MIMUW} \and Karol
Wegrzycki\footrecall{MIMUW} \and Adam Witkowski\footrecall{MIMUW}\and Piotr Wygocki\footrecall{MIMUW}}
\date{\texttt{[apacuk,sank,k.wegrzycki,a.witkowski,wygos]@mimuw.edu.pl}}


\maketitle
\input{macros}
\input{abstract}


\input{challenge_description}

\input{solution}
\input{conclusion}

%
\bibliographystyle{abbrv}
\bibliography{bib}  

\end{document}

%% file: macros.tex
\newcommand{\GT}{\ensuremath{G}\xspace}
\newcommand{\target}{\ensuremath{T}\xspace}
\newcommand{\items}{\ensuremath{I}\xspace}
\newcommand{\score}{\ensuremath{\mathsf{score}}\xspace}
\newcommand{\scorearg}[2]{\score \ensuremath{(#1, #2)}\xspace}
\newcommand{\interactions}[1]{\ensuremath{\mathsf{Int}_{#1}}\xspace}
\newcommand{\deletes}[1]{\ensuremath{\mathsf{Del}_{#1}}\xspace}
\newcommand{\impressions}[1]{\ensuremath{\mathsf{Imp}_{#1}}\xspace}
\newcommand{\similar}[2]{\ensuremath{\mathsf{sim}({#1}, {#2})}\xspace}
\newcommand{\similarnoargs}{\ensuremath{\mathsf{sim}}\xspace}
\newcommand{\intsimilar}[2]{\interactions{}\ensuremath{\text{-}}\similar{#1}{#2}}
\newcommand{\intsimilarnoargs}{\interactions{}\ensuremath{\text{-}}\similarnoargs}
\newcommand{\impsimilar}[2]{\impressions{}\ensuremath{\text{-}}\similar{#1}{#2}}
\newcommand{\impsimilarnoargs}{\mpressions{}\ensuremath{\text{-}}\similarnoargs}
\newcommand{\itemsimilar}[2]{\items\ensuremath{\text{-}}\similar{#1}{#2}}
\newcommand{\itemsimilarnoargs}{\items\ensuremath{\text{-}}\similarnoargs}
\newcommand{\jaccard}[2]{\ensuremath{J({#1},{#2})}\xspace}

\newcommand{\intusers}[1]{\ensuremath{\interactions{}\text{-}\mathsf{users}(#1)}\xspace}
\newcommand{\probab}[1]{\ensuremath{\mathbb{P}\left[ #1 \right]}\xspace}
\newcommand{\pred}{\ensuremath{\mathsf{pred}}\xspace}
\newcommand{\predarg}[1]{\pred \ensuremath{(#1)}\xspace}

\newcommand{\intersect}[2]{\ensuremath{|{#1} \cap {#2}|}\xspace}
\newcommand{\commontags}[2]{\ensuremath{\mathsf{common\text{-}tags(#1, #2)}}\xspace}
\newcommand{\commontitle}[2]{\ensuremath{\mathsf{common\text{-}title(#1, #2)}}\xspace}
\newcommand{\tags}[1]{\ensuremath{\mathsf{tags}(#1)}\xspace}
\newcommand{\titles}[1]{\ensuremath{\mathsf{title}(#1)}\xspace}
\newcommand{\jobroles}[1]{\ensuremath{\mathsf{jobroles}(#1)}\xspace}

\newcommand{\cluster}[1]{\ensuremath{\mathsf{cluster}(#1)}\xspace}

%% file: abstract.tex
\begin{abstract}

We present the Mim-Solution's approach to the RecSys Challenge 2016, which ranked 2nd. 
The goal of the competition was to prepare job
recommendations for the users of the website Xing.com.

Our two phase algorithm consists of candidate selection followed by the
candidate ranking.  We ranked the candidates by the predicted probability that
the user will positively interact with the job offer. We have used Gradient
Boosting Decision Trees as the regression tool.

\end{abstract}

%% file: challenge_description.tex
\section{Introduction}\label{recsys_desc}

The Recsys Challenge is an annual competition of recommender systems.  The
2016th edition\footnote{\url{http://2016.recsyschallenge.com/}} was based on data
provided by \url{xing.com} -- a platform for business networking.  On Xing, users
search for job offers that could fit them.  Each user has a profile containing
information such as: place of living, industry branch and experience (in years). Job
offers are described by a related set of properties.  There are various ways in
which a user can interact with a job offer (called item): by clicking is, as well as bookmarking
interesting ones, replying to offers and finally by deleting a recommendation.

The task in the challenge was to predict for a given XING user  30 items
that this user will positively interact with (click, bookmark or reply to).

\paragraph{Dataset}\label{dataset}

The dataset consisted of properties of users and items, interactions of the users
(which items user clicked, bookmarked, replied to or deleted) and impressions (items shown to
users by XING recommender system).
The interactions and impressions were gathered from a 3 month period.

All data was anonymized by changing all properties to numerical values
and adding an unknown number of artificial users, items, interactions and impressions.

We were also given a set of both user and item properties. Common attributes
of users and items were: career level, discipline id, industry id, country and
region.  Besides that users had following attributes: jobroles, experience n
entries class, experience years experience, experience years in current, edu
degree, edu fieldofstudies. Items had attributes: title, latitude,
longitude, employment, created at and active during test\footnote{
A more detailed specification of the dataset can be found on
\url{https://github.com/recsyschallenge/2016/blob/master/TrainingDataset.md}}.
Values of those attributes we will denote by, e.g., $\mathsf{career\_level}(i)$ for a given item $i$.

Each impression is a tuple containing: user ID, item ID, number of the week in which
impression occurred. Each interaction contains: user ID, item ID,
interaction type (click, bookmark, reply, delete) and timestamp of the
interaction.
We will denote the positive interactions of a user $u$ by \interactions{u},
the negative interactions by \deletes{u} and impressions by \impressions{u}.

$10$\% of users were target users: users which we needed to compute
the predictions for. The predicted items had to come from a subset ($24$\%) of all items
(these were the job offers open during the period).
Exact datasets sizes are presented in Table~\ref{tab:data_size}.

\begin{table}[ht!]
    \centering
    \caption{Statistics of the Recsys Challenge 2016 training dataset.}
    \begin{tabular}{lr}
        dataset                             & size                          \\
        \hline
        impressions                         & $1\,078\,627\,301$            \\
        interactions                        & $8\,826\,678$                 \\
        users (target users)                & $1\,500\,000$ ($150\,000$)    \\
        items (items active during test)    & $1\,358\,098$ ($327\,003$)    \\
    \end{tabular}
\label{tab:data_size}
\end{table}

\paragraph{Evaluation measure}\label{score_sec}
The \emph{ground truth} is a mapping that assigns to each user,
the set of items he interacted positively with during the test week.
Let \target be the set of target users, \items be the set of all items,
and $\GT : \target \rightarrow  2^{\items}$ the ground truth.
We will also denote the sequence of items predicted for a user by \predarg{u}.

The quality of recommendations was measured by a function:
\begin{equation}\label{score_def}
    \score(\pred, G) = \sum_{u \in T} \mathsf{userScore}(\predarg{u}, G(u)),
\end{equation}
where:
\begin{align*}
\mathsf{userScore}(\predarg{u}, \GT (u)) = \\
 20\cdot \big[ \mathsf{p}(\predarg{u}, \GT (u), 2) &+ \mathsf{p}(\predarg{u}, \GT (u), 4) + \\
   \mathsf{us}(\predarg{u}, \GT (u)) &+ \mathsf{r}(\predarg{u}, \GT (u)) \big] + \\
 10 \cdot \big[ \mathsf{p}(\predarg{u}, \GT (u), 6) &+ \mathsf{p}(\predarg{u}, \GT (u), 20) \big] \\
\end{align*}
and for a sequence $\bar{a} = a_1,a_2, \dots, a_{30}$, a set $B$ and a natural number $k$:
\begin{align*}
\mathsf{p}(\bar{a},B,k) &= |\{a_1,a_2,\ldots,a_k\} \cap B| / k \\
\mathsf{r}(\bar{a}, B) &= |\{a_1,a_2, \ldots,a_{30}\} \cap B| / \min(1,|B|) \\
\mathsf{us}(\bar{a}, B) &= \min(1, |\{a_1,a_2, \ldots,a_{30}\} \cap B|).
\end{align*}
Note that:
$\mathsf{p}(\bar{a}, B, k)$ is the precision for the first
$k$ elements of $a$, $\mathsf{r}(\bar{a}, B)$
is the recall and $\mathsf{us}(\bar{a}, B)$ is user success (1 if if we predicted at least one item for this user correctly, 0 otherwise).

Solutions were evaluated by the submission system and contestants received
instant feedback with \score{} value calculated on a sample of $\frac{1}{3}$ of target users.

%% file: solution.tex
\section{Our solution}\label{solution}


Our solution consists of two parts:
\begin{enumerate}
 \setlength\itemsep{0em}
 \item for each user we calculate a set of candidate items, much smaller than the whole set \items{} (Only those candidates were considered when creating a submission),
\item learn for each (user, candidate) pair $(u,i)$ the probability $\probab{i \in \GT(u)}$
that the user will interact with this item.
\end{enumerate}

We submitted, for each user, 30 items with the highest predicted probability of interaction,
excluding items that the user deleted.
Limiting the number of considered items per user allowed us to substantially
decrease the time required to train a model and prepare a submission.
Examining all possible user-item pairs ($150.000 \times 327.000$ pairs) was
infeasible considering our resources.  To learn the probabilities $\probab{i
\in \GT(u)}$, we have used Gradient Boosting Decision Trees (GBDT)
\cite{DBLP:journals/corr/ChenG16}\footnote{\url{https://github.com/dmlc/xgboost}},
optimizing the logloss measure. We learned the probabilities instead of
directly optimizing the \score{} function since it does not give results for a single user-item pair.
The schema of our solution is presented in Figure~\ref{fig:architecture}.
\input{architecture}

Our solution was ranked 2nd in the competition,
scoring 675985.03 and 2035964.16 points on the public and private leaderboard respectively.
To put this into perspective, submitting for each user $u$, sorted from most recent \interactions{u} but without items from \deletes{u}
(adding \impressions{u} if there were less than 30 unique interactions)
achieved a score of 495k on the public leaderboard.
All of the computation were performed on 12 cores (24 threads), 64 GB RAM Linux server.




\subsection{Training set}\label{val_set}
In this problem, there was no clearly defined training set and the first challenge was to create it.
Since our task was to predict users' interactions in the week following the
end of the available data, we trained our model on all the data except the last available week and then
used this last week data to compute the training ground truth \GT{} to evaluate the model.
This way, we could calculate the score and determine if we are making progress
without sending an official submission (every team was allowed max. 5 submissions per day).

There was an overlap between the training data and
the test data (the full dataset). Both candidates and features were computed
separately for the training set and the full dataset.


\subsection{Candidate items selection}\label{cands}
Since there were 150k target users and more than 300k items,
making a prediction for each user-item pair would take too long.
To address this issue, we chose for each user $u$ a set of promising items, which we called candidates.

To define candidates, we will need a few notions of similarity.
The Jaccard coefficient between two sets $A$ and $B$ is $\jaccard{A}{B} \equiv \frac{|A \cap B|}{|A \cup B|}$.
Interactions (impressions) similarity between two users $u, u'$, denoted $\intsimilar{u}{u'}$ ($\impsimilar{u}{u'}$),
is the Jaccard coefficient between the sets of items from their positive interactions (impressions).
For example $\intsimilar{u}{u'} = \jaccard{\interactions{u}}{\interactions{u'}}$.

For items $i, i'$, we will denote $\commontags{i}{i'} \equiv |\tags{i} \cap \tags{i'}|$
and $\commontitle{i}{i'} \equiv |\titles{i} \cap \titles{i'}|$.

For an user $u$, the candidates were:
\begin{enumerate}
 \setlength\itemsep{0em}
 \item \label{interactions} $\interactions{u}$ sorted by the week of occurrence (most recent first) and the number of interactions,
 \item \label{impressions} $\impressions{u}$ sorted the same way as \ref{interactions},
 \item \label{interaction-similar} $\interactions{u'}$ for users $u'$ with large $\intsimilar{u}{u'}$,
 \item \label{impression-similar} $\impressions{u'}$ of users $u'$ with large $\impsimilar{u}{u'}$,
 \item \label{knn-interactions} items $i$ with large $\max_{i' \in \interactions{u}} \commontags{i}{i'}$,
 similarly for $\max_{i' \in \interactions{u}} \commontitle{i}{i'}$, $\max \intersect{\titles{i'}}{\tags{i}}$
and $\max \intersect{\titles{i}}{\tags{i'}}$,
 \item same as \ref{knn-interactions}, just with $\max$ taken over $i' \in \impressions{u}$,
 \item items $i$ with large \intersect{\jobroles{u}}{\tags{i}},
 \item items $i$ with large \intersect{\jobroles{u}}{\titles{i}},
 \item the most popular items (globally, this list was the same for all users).
\end{enumerate}
The popularity of an item was measured by the number of interactions of all users with this item.
From each category, we took 60 best candidates for each user.
This approach gave us 43M (user,item) pairs, on average just short of 300 candidates per user.
On the training set, candidates chosen this way covered 37\% of the training ground truth.

Of course, we could have chosen a different notions of similarity between users/items.
In particular we considered similarity based on user and item properties such as region, industry, etc.
Candidates based on those measures of similarity did not improve the score sufficiently, probably due to anonymization of the data.




\subsection{Learning the probabilities}\label{xgboost}
We wanted to construct a model that given a $(u,i)$ pair
and values of features for this pair will compute the probability
$\probab{i \in \GT(u)}$.
In order to estimate the probabilities, we have used XGBoost\footnote{\url{https://github.com/dmlc/xgboost}},
a machine learning library implementing GBDT.

For training the ranking model, we split all the users with at least one item
in the training ground truth into two sets of equal size.
The users from the first set were used for training XGBoost model,
and the users from the second set were used for validation of the model.
The training file (and validation file) contained for each user:
\begin{itemize}
 \setlength\itemsep{0em}
    \item all the training candidate items, which occurred in the training ground truth (positive candidates)
    \item and up to $5$ training candidate items, which did not occur in the training ground truth (negative candidates).
\end{itemize}

Just after deadline for submitting solutions
we observed, that training a model on all users with all positive and $1/4$ of
negative candidates improves our score by 6.5k points on unofficial public leaderboard\footnote{
Note that difference between 1st and 2nd place was 5.7k, so extending training
file earlier could result in winning a competition by our team.
}.

\paragraph{Evaluation}
We used maximum likelihood as the objective function optimized by GBDT.
Additionally we verified models by computing
the \score{} function based on the validation part of the training ground truth.
Major improvements in this validation \score{} translated to comparable
improvements on the score achieved via the submission system.
We measured on training ground truth that our way of ordering previously selected candidate items
achieved score which was 77.5\% of best possible result based only on those candidates.


\paragraph{Parameters tuning}
We found that the optimal XGBoost parameters for our task were:
\begin{itemize}
 \setlength\itemsep{0em}
    \item maximum depth of a tree (max\_depth) in range $[4,6]$,
    \item minimum weight of node to be splitted (min\_child\_weight) in range $[4,6]$,
    \item learning rate (eta) $=0.1$,
    \item minimum loss reduction to make a node partition (gamma) $=1.0$,
    \item number of rounds (num\_round) $=1000$ (validation logloss did not improve after $1000$ rounds).
\end{itemize}
%
%
\subsection{Features}\label{feats}
Each feature is a function that for user $u$ and item $i$ maps the pair $(u,i)$ to some real number.
We ended up with 273 features. Many of these features were highly correlated,
however, we observed that redundant features do not reduce the quality of the model.
Many features differ only in:
\begin{itemize}
 \setlength\itemsep{0em}
    \item used different events source: impressions instead of interactions, only positive/negative interactions, only interactions of one type,
    \item instead of Jaccard coefficient we used size of sets intersection $|A \cap B|$ and vice-versa,
    \item used various aggregate functions: maximum, minimum, sum, average, count or unique count (count without duplicate entries).
\end{itemize}
Because of this we will only describe the important groups of features. There are 12 such groups
Table~\ref{tab:feature_importance} summarizes the importance of the features we used.
Ideas for features were inspired by papers of previous RecSys Challenge~\cite{Romov:2015:RCE:2813448.2813510,Volkovs:2015:TAI:2813448.2813512}
and similar competitions hosted on Kaggle platform\footnote{\url{www.kaggle.com}}.

\begin{table}[ht!]
    \centering
    \caption{Feature groups importance: sum of how many times
             each feature was used to split the data (based on XGBoost's model fscore metric).
             The main features groups are in bold, additionally we list important subgroups.}
    \begin{tabular}{|lr|}
	 \hline
        feature group                                   & importance    \\
        \hline
        \textbf{event based}                            & \textbf{5499} \\
        tags + title                                    & 946           \\
        \textbf{item global popularity}                 & \textbf{2913} \\
        trend                                           & 1392          \\
        weekday                                         & 599           \\
        \textbf{cf most similar}                        & \textbf{1410} \\
        item clicked by user                            & 790           \\
        user who clicked item                           & 620           \\
        \textbf{user total events}                      & \textbf{1066} \\
        in last week                                    & 494           \\
        \textbf{seconds from last user activity}        & \textbf{882}  \\
        \textbf{max common tags with clicked item}      & \textbf{527}  \\
        \textbf{position on the candidates list}        & \textbf{456}  \\
        \textbf{user-item events count in last week}    & \textbf{288}  \\
        \textbf{item properties}                        & \textbf{375}  \\
        created at                                      & 166           \\
        longitude                                       & 125           \\
        latitude                                        & 84            \\
        \textbf{content based user-item similarity}     & \textbf{83}   \\
        career level difference                         & 47            \\
        count of common user job roles and item titles  & 15            \\
        \textbf{distance to the closest clicked item}   & \textbf{55}   \\
        \textbf{items cluster}                          & \textbf{7}    \\
	\hline
    \end{tabular}
\label{tab:feature_importance}
\end{table}

\noindent \textbf{Event based} features are
percentages of items
from $\interactions{u}$ that had some property (e.g., item's career level) equal to
item's $i$ corresponding property.
Dually we also used the percentages of users from $\intusers{i}$ (i.e., users who interact a given item $i$) that had
some attribute equal to user $u$ attribute.
From this group of features, the best were those based
on item tags and item title fields.

\noindent\textbf{Item global popularity} is
the number of times item $i$ was clicked by any user.
Additionally we computed the trend of popularity: clicks count in the last week divided by
the clicks count in the previous last week.
Another way to observe week trend was to
compute trend based on events from last and previous last Mondays, Tuesdays,
\ldots, Sundays.

\noindent\textbf{Colaborative filtering most similar} are features that measure similarity
between the item $i$ and the items from \interactions{u} using \itemsimilarnoargs,
and dually between the user $u$ and users that interacted with $i$, using \intsimilarnoargs.
Formally, these are given by formulas:
\begin{itemize}
 \setlength\itemsep{0em}
\item $\max_{i' \in \interactions{u}} \itemsimilar{i}{i'}$,
\item $\max_{u' \in \intusers{i}} \intsimilar{u}{u'}$.
\end{itemize}

\noindent\textbf{User total events} is just $|\interactions{u}|$
both with repetitions of items and without.
We also used this feature limited to the user's last week of activity.

\noindent\textbf{Seconds from last user activity} are the differences, in seconds, between:
\begin{itemize}
 \setlength\itemsep{0em}
\item the last time $u$ clicked $i$ and the maximal timestamp in data,
\item the last time $u$ clicked $i$ and the last time $u$ clicked any item,
\item the last time $u$ clicked any item and the maximal timestamp in data.
\end{itemize}
Analogous features based on impressions were timed in weeks.

\noindent \textbf{Max common tags with clicked item} are features
that return the maximum number of common tags  between item $i$
and items from \interactions{u}: $\max_{i' \in \interactions{u}} \commontags{i}{i'}$.
We used those features with \commontitle{i}{i'} instead of \commontags{i}{i'} and
with impressions instead of interactions.

\noindent \textbf{Position on the candidates list} is
position number of item $i$ on user's $u$ candidates list.
There was separate feature for each candidate algorithm (see Subsection \ref{cands}).

\noindent \textbf{User-item events count in last week} is the count of
 how many times user $u$ clicked item $i$ in the last week of this users activity,
or in the last week from the dataset.

\noindent \textbf{Item properties} are values of item's $i$ attributes.

\noindent \textbf{Content based user-item similarity} is a group of features based only on
properties of $u$ and $i$. The features were:
\begin{itemize}
 \setlength\itemsep{0em}
\item $\mathsf{career\_level}(i) - \mathsf{career\_level}(u)$,
\item \intersect{\jobroles{u}}{\titles{i}},
\item \intersect{\jobroles{u}}{\tags{i}}.
\item $1$ if $\mathsf{attr}(u)=\mathsf{attr}(i)$ else $0$,
      for the rest of matching attributes.
\end{itemize}

\noindent\textbf{Distance to the closest clicked item} is the Euclidean distance between the location of item $i$
and the location of the closest item from \interactions{u}.

\noindent\textbf{Item cluster} is a boolean feature, true if the item $i$
is in a cluster of some item from \interactions{u},
where $i \in \cluster{i'}$ if $i \neq i'$
and there exists a user $u$ that clicked both $i$ and $i'$ within 10 minutes.
This feature is another variation of the similarity between items.
This time, we say that two items are similar if a user clicked both of them
in a similar time. The motivation is simple: if both those items were interesting for some user,
then they probably have something in common that appeals to this user.

Some of the features (especially time-related) had to be calculated separately and with care for the training dataset and the full dataset.
For example, we had a feature ``timestamp of the last interaction''. This feature on training instance should be shifted in order to cover the same range
of values as on the full dataset.

\subsection{Blending}\label{blending}
In a final step in the construction of our solution we merged
our best models.
Since all of them were similar and XGBoost based, we took for each $(u,i)$ pair,
the arithmetic mean of probabilities  $\probab{i \in \GT(u)}$ calculated by those models.
Finally, for each target user, we sorted candidate items by
these averaged probabilities and selected the top 30.

%% file: architecture.tex
\newcommand{\jobnode}[3]{
\node[draw, rounded corners=.05cm] (item) at (#2,#3) {job \##1};
}

\newcommand{\probnode}[4]{
\node[rectangle split, rectangle split parts=2, draw, rounded corners=.05cm] 
at (#3,#4)  { job \##1 \nodepart{two} #2};
}

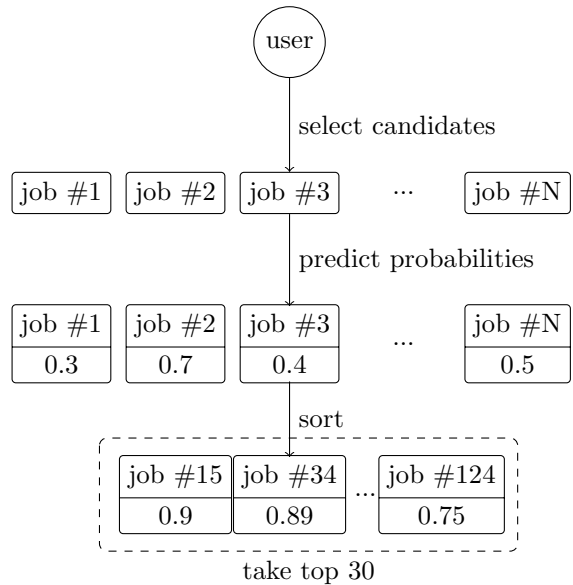
\begin{figure}[!ht]
    \begin{center}
\begin{tikzpicture}
 \node[draw, circle] (u) at (3,3) {user};
 \jobnode{1}{0}{1}
 \jobnode{2}{1.5}{1}
 \jobnode{3}{3}{1}
 \draw[->] (u) -- (item) node[midway, right] {select candidates};
 \draw[->] (item) -- (3,-0.5) node[midway, right] {predict probabilities};
 \draw[->] (3,-1.5) -- (3,-2.5) node[midway, right] {sort}; 
 \node at (4.5,1) {...};
 \jobnode{N}{6}{1}
 \probnode{1}{0.3}{0}{-1}
 \probnode{2}{0.7}{1.5}{-1}
 \probnode{3}{0.4}{3}{-1}
 \node at (4.5,-1) {...};
 \probnode{N}{0.5}{6}{-1}
 \probnode{15}{0.9}{1.5}{-3}
 \probnode{34}{0.89}{3}{-3}
 \node at (4,-3) {...};
 \probnode{124}{0.75}{5}{-3}
 \draw[rounded corners=.1cm, dashed] (0.5, -2.25) rectangle (6,-3.75);
 \path (0.5,-3.75) -- (6,-3.75) node[midway, below] {take top 30};
\end{tikzpicture}
\caption{Schema of our solution.}
\label{fig:architecture}
    \end{center}
\end{figure}

%% file: conclusion.tex
\section{Conclusions}\label{conclusion}
We have presented Mim-Solution's approach to the 2016th RecSys Challenge.
We used XGBoost to predict probabilities that a user will be interested in a job offer,
but only for preselected offers. This allowed to efficiently train and evaluate the model as well as use
complex and robust features.

Even tough our score was good (28.5 \% of best possible score measured on our train data),
there was definitely a lot of room to improve.
One easy improvement, mentioned already in the paper, is increasing the size of the training set.
There was also some place for improvements in ordering of candidate items,
but we suspect that twice bigger room for improvement was in expanding the set of candidates
(we achieved 77.5\% and 37\% of best possible results in the layer of sorting and selecting candidates respectively).

This work is supported by ERC project PAAl-POC 680912.